\documentclass[12pt]{article}

\usepackage{newtxtext,newtxmath}

\usepackage{graphicx}

\usepackage[letterpaper,margin=1in]{geometry}

\renewenvironment{abstract}
	{\quotation}
	{\endquotation}

\date{}

\makeatletter
\renewcommand{\fnum@figure}{\textbf{Figure \thefigure}}
\renewcommand{\fnum@table}{\textbf{Table \thetable}}
\makeatother

\usepackage{scicite}

\usepackage{url}

\def\scititle{
	Mining beyond Earth with Space Robots: Exploration, Sampling, and Extraction
}
\title{\bfseries \boldmath \scititle}

\author{
Dong Li$^{1,11}$, Dujun Nie$^{1,11}$, Xiaotong Zhang$^{1,11}$, Ruilin Wang$^{1,11}$, Yuchen Li$^{2,3}$\and Chang Ge$^{4}$, Chao Xiong$^{5}$, Kaichang Di$^{6}$, Andreas Nüchter$^{7}$, Levente Kovács$^{3}$\and  Qingquan Li$^{8}$, Shirong Ge$^{9}$, Fei-Yue Wang$^{1}$, Long Chen$^{1,10,11,*}$ \and
	\small$^{1}$Institute of Automation, Chinese Academy of Sciences, Beijing, 100190, China\and
	\small$^{2}$School of Computation, Information and Technology, Technical University of Munich, Munich, 80333, Germany\and
    \small$^{3}$University Research and Innovation Center, Obuda University, Budapest, 1034, Hungary\and
    \small$^{4}$School of Mechanical Engineering and Automation, Beihang University, Beijing, 100191, China\and
    \small$^{5}$School of Earth and Space Sciences, Wuhan University, Wuhan, 430072, China\and
    \small$^{6}$State Key Laboratory of Remote Sensing Science, Aerospace Information Research Institute, \and
    \small Chinese Academy of Sciences, Beijing, 100190, China\and
    \small$^{7}$Faculty of Mathematics and Computer Science, University of Würzburg, Würzburg, 97074, Germany\and
    \small$^{8}$School of Architecture \& Urban Planning, Shenzhen University, Shenzhen, 518060, China\and
    \small$^{9}$School of Mechanical and Electrical Engineering, China University of Mining and Technology-Beijing\and 
    \small Beijing, 100083, China\and
    \small$^{10}$Institude of Research \& Development, WAYTOUS, Beijing, 100190, China\and
    \small$^{11}$OpenSpace Lab, China\and
	\small$^\ast$Corresponding author. Email: long.chen@ia.ac.cn
}

\begin{document} 

\maketitle

\begin{abstract}

Space resource acquisition and utilization, commonly referred to as Space Mining, represent critical pathways for enabling sustained human exploration and unlocking commercial opportunities in space. These resources mainly include helium-3, water, mineral resources on the Moon and Mars, and abundant mineral deposits on asteroids. Due to the harsh conditions of space, communication delays, and high launch costs, the development of autonomous robotic systems is critical to achieving efficient, cost-effective space mining. This paper provides a comprehensive overview of space mining robotics and associated technologies. First, we review the background of space mining, including international policies, commercial entities, and recent advancements. We define a systematic six-stage architecture for space mining: \textbf{Exploration} is initiated by (1) remote sensing for target identification and (2) precise in situ robotic detection; \textbf{Sampling} progresses from (3) single-robot small-scale sampling to (4) multi-robot large-scale excavation; and \textbf{Extraction} integrates (5) autonomous resource extraction and (6) final integration into in situ construction or terrestrial transport. Additionally, we review and curate existing resources for space mining research, including real-world mission data, terrestrial analog datasets, and high-fidelity simulation environments. Finally, we identify critical open challenges in autonomous space mining and delineate a strategic research roadmap to bridge current technological gaps, fostering the transition toward a sustainable off-world economy. To track ongoing developments in space mining, we maintain an updated project page: https://github.com/OpenSpace-Lab/Space-Mining-with-Robotics-List.

\end{abstract}

\section{Introduction}

\begin{figure*}[t]
    \centering
    \includegraphics[width=1\textwidth]{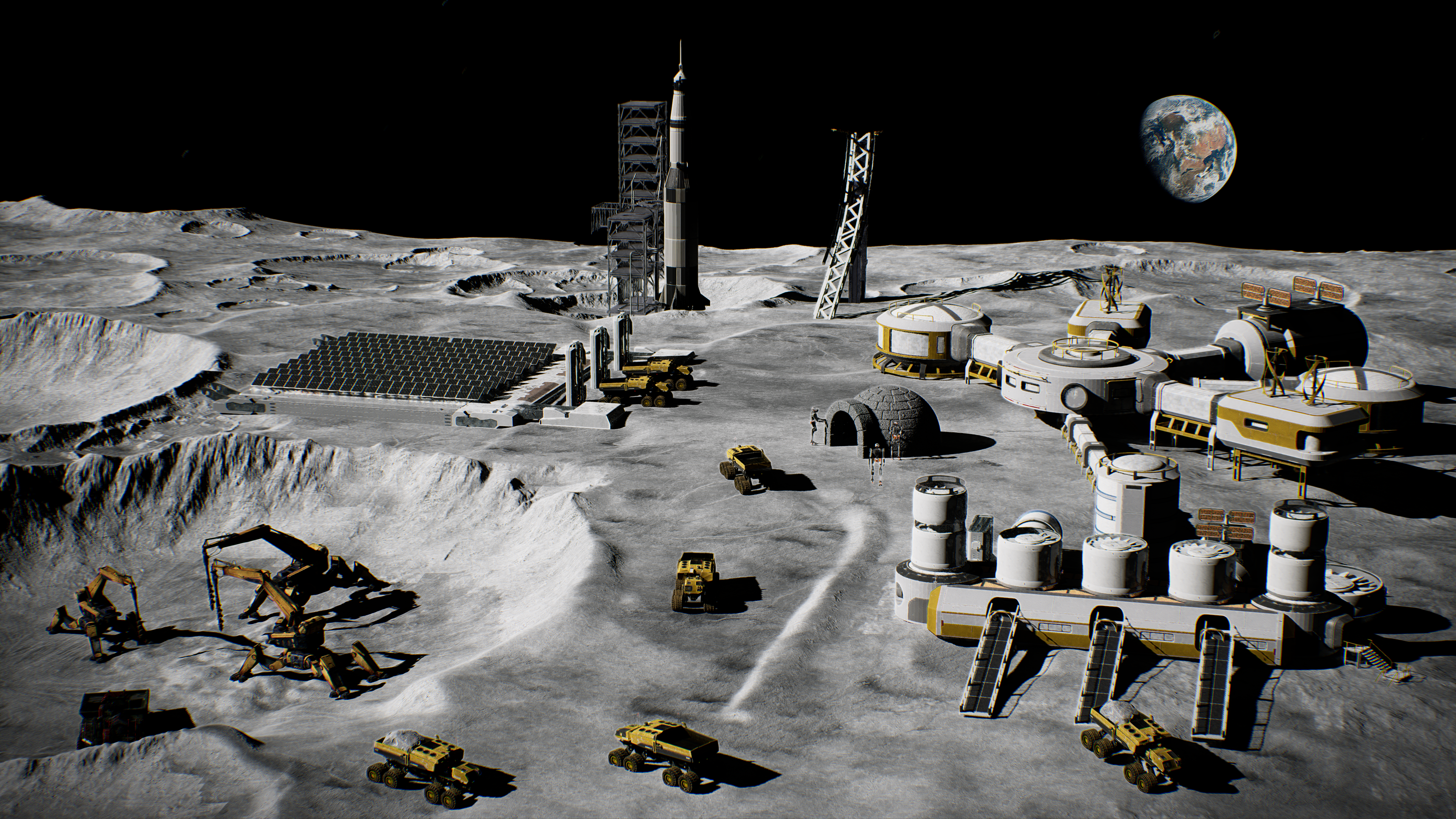}
    \vspace{-25pt} 
    \caption{\textbf{Conceptual Overview of Space Mining.} Schematic of the multi-stage process for 
extraterrestrial resource extraction and utilization, from prospecting and excavation, through transport 
to processing facilities or depots, to in-situ or orbital refining of volatiles, metals, and regolith, 
culminating in applications for propellant production, life support, construction, or return to Earth. (Image credit: Original work by the authors.)  }
    \label{fig:comparison}
\end{figure*}

The rapid depletion of Earth’s finite mineral reserves driven by surging industrial demand presents a foundational challenge to long-term sustainability. As high-grade terrestrial deposits of critical materials approach exhaustion, the quest for extraterrestrial supply chains has transcended theoretical discourse to become a geopolitical and economic urgency~\cite{hein2013deep}. This transition from a single-planet species to a spacefaring civilization hinges upon the mastery of In-Situ Resource Utilization (ISRU), a framework that facilitates the emerging industry of space mining~\cite{starr2020mars}. By tapping into the vast mineralogical potential of the Moon, Mars, and near-Earth asteroids, humanity can decouple mission architectures from Earth’s gravity well and bypass the physical constraints of a single-planet economy, and accelerate the construction of extraterrestrial infrastructure.

The strategic exploitation of extraterrestrial resources serves two interconnected imperatives, both ultimately converging toward the long-term sustainability of human civilization in the Solar System. First, the extraction of lunar water ice enables life-support systems, while the harvesting of helium-3 (³He) provides high-energy-density fusion fuel; together, these resources form the essential material foundation for sustained deep-space operations~\cite{fegley1993lunar,colaprete2010detection}. Second, the mineral-rich compositions of near-Earth asteroids offer not only strategic raw materials for terrestrial industry, but also a viable alternative to conventional mining, thereby alleviating environmental pressures on Earth~\cite{michel2025hera}. The critical enabler bridging these two objectives lies in the deep integration of robotic autonomy with space resource acquisition, a leap that represents the decisive step toward securing long-term human prosperity across the Solar System.

With swift advances in U.S. and Chinese commercial spaceflight and exploration, alongside falling launch and payload costs, lunar exploration has shifted from symbolic milestone to pragmatic race for resource sovereignty. Although Artemis II's 2026 circumlunar mission marks a pivotal transition, and other players like Russia, Japan, and the EU actively participate, the global landscape remains defined by U.S.–China strategic competition. As these superpowers converge on the volatile-rich lunar south pole, resource extraction has become the crux of a multi-polar industrial race. Yet significant technical and logistical bottlenecks in autonomous perception and robust manipulation still render large-scale space mining a formidable challenge.

Although the deployment of intelligent robots and autonomous vehicles for large-scale mineral resource extraction is well-established on Earth
~\cite{chen2024sustainable,chen2024autonomous,ge2022making,roberts2002reactive}, significant challenges arise when transitioning these applications to extraterrestrial environments, such as the Moon, planets, and asteroids~\cite{xu2020approach,dallas2020mining}. 
The main challenges of space mining can be categorized as follows:
(1). \textit{Economic and Logistical Constraints}: The primary barrier to large-scale space mining is the exorbitant cost of launching payloads into orbit, which necessitates a paradigm shift toward high-efficiency, lightweight robotic designs~\cite{gao2017review}. Beyond initial launch costs, the logistical complexity of transporting heavy mining infrastructure across interplanetary distances imposes strict mass and volume constraints on automated equipment. These economic pressures drive the demand for modularity and the integration of In-Situ Resource Utilization (ISRU) to minimize dependence on Earth-sourced supplies, thereby ensuring the long-term financial viability of extra-terrestrial missions~\cite{sonter1997technical,bowkett2025autonomous}.
(2). \textit{Environmental Constraints}: The primary characteristic of extraterrestrial environments is the microgravity condition, which presents substantial challenges for the design of robotic systems and automated equipment, as well as for the development of effective mining methodologies~\cite{jiang2017robotic,colaprete2019overview}. Furthermore, environments such as those on the Moon and Mars introduce additional complexities, including pervasive dust, sandstorms, extreme temperature fluctuations, and elevated radiation levels~\cite{pedersen2003survey,gao2017review}.
(3). \textit{Resource Extraction and Processing}: Extraterrestrial resources are found on the Moon, Mars, other planets, and near-Earth asteroids, each varying significantly in size, resource type, and mode of occurrence, which presents substantial challenges to extraction methodologies~\cite{vergaaij2021economic,metzger2021aqua}. Additionally, certain resources present high extraction difficulties, such as mining metallic asteroids and transporting materials back to Earth, as well as extracting $^3$He and water ice from lunar regolith for ISRU. These complexities require the development of specialized robotic systems and adaptive processing techniques to enable efficient and sustainable resource extraction in various extraterrestrial environments~\cite{schuler2022isru}.
(4). \textit{Communication and Autonomy}: Due to the vast distances involved, robotic and automated systems for space mining face significant communication delays, ranging from seconds for lunar operations to minutes for Martian missions~\cite{mcbrayer2022communication}. Consequently, the development of highly autonomous robots and automated equipment capable of independent decision-making and robust fault recovery is essential to achieving efficient and reliable space mining operations.
(5). \textit{Scarcity of Testing Environments}: The high cost of space-based experimentation significantly hinders the development and validation of robust space mining robots. Limited access to realistic testing environments that replicate extraterrestrial conditions, such as microgravity, lunar or Martian regolith, and extreme radiation, poses a major challenge in demonstrating the feasibility and reliability of robotic systems for space mining applications~\cite{yang2018grand}.
As summarized in Table~\ref{tab:env-compare}, the vast disparities in gravity, thermal cycles, and communication latency across these celestial targets necessitate adaptive operational frameworks to overcome the fundamental hurdles of autonomous space mining.

\begin{table*}[ht]
\centering
\caption{Comparative environmental and operational conditions for lunar, Martian, and near-Earth asteroid (NEA) missions.}
\vspace{10pt}
\label{tab:env-compare}
\scriptsize  
\begin{tabular}{l c c c}
\hline
\textbf{Parameter} & \textbf{Moon} & \textbf{Mars} & \textbf{Near-Earth Asteroid} \\
\hline
\textbf{Temperature Range} & $-$190°C -- 127°C & $-$143°C -- 35°C & $\sim-$270°C \\
\textbf{Surface Gravity} & 1.625 m/s² (1/6 $g$) & 3.71 m/s² (0.38 $g$) & Microgravity ($10^{-6}$ to $10^{-3}$ m/s²) \\
\textbf{Day-Night Cycle} & $\sim$29 Earth days & 24 h 39 min & N/A (rotational period varies) \\
\textbf{Atmosphere} & None & Thin & None \\
\textbf{Communication Delay} & $\sim$1.3 s (one-way) & 4--24 min (one-way) & 10 s to $\sim$20 min (varies by orbit) \\
\textbf{Key Resources} & $^3$He, water ice, rare earths & Water (ice/polar), metals (in ores) & Metals (Ni-Fe), volatiles \\
\textbf{Notable Hazards} & 
  High solar radiation, dust &
  Dust storms, CO$_2$ sublimation &
  Irregular shape, spin, low escape velocity \\
\textbf{Mapping Maturity} & High (LRO, Chang'e) & High (MRO, orbiters $>10$ yrs) & Low (varies by target) \\
\hline
\end{tabular}
\end{table*}

\textbf{Scope.} This paper examines the critical intersection of space mining and autonomous robotic systems. First, we provide a comprehensive synthesis of the extraterrestrial resource landscape, evaluating recent aerospace milestones, regulatory frameworks, and industrial ecosystems. Second, we propose a systematic six-stage operational framework for space mining, which serves as the foundation for a rigorous feasibility analysis of current robotic technologies across structural design, perception, autonomous planning, and system-level execution. Third, we curate an essential repository of validation resources, including multi-source mission datasets, high-fidelity simulation environments, and physical testbeds. Finally, we identify the primary technical bottlenecks and delineate a strategic roadmap for the future, highlighting the transformative role of advanced algorithms and resilient hardware in realizing autonomous off-world resource acquisition. A conceptual schematic of the space mining pipeline is illustrated in Fig.~\ref{fig:comparison}.
The main contributions of this survey are summarized as follows:
\begin{itemize}
    \item \textbf{Detailed overview of space mining background:} We provide a comprehensive review of the current space mining landscape, surveying recent aerospace advancements, robotic deployments, and the evolving commercial-legal infrastructure. By drawing parallels with established terrestrial open-pit operations, we demonstrate that Earth-based mining paradigms provide a robust technological and operational foundation for space mining, while delineating the unique divergences necessitated by the space environment.
    \item \textbf{A systematic synthesis of space mining technologies through a six-stage framework:} We propose a hierarchical six-stage architecture for space mining, establishing a seamless operational continuum across three core phases: Exploration, Sampling, and Extraction. This framework delineates the systematic progression from initial reconnaissance to autonomous extraction and final implementation, providing a comprehensive roadmap for the entire extraterrestrial resource value chain.
    \item \textbf{Curation of high-fidelity research infrastructures:} Recognizing that exorbitant launch costs and limited mission frequency severely constrain data availability, we curate a comprehensive repository to bridge this gap. This resource integrates multi-source telemetry from extra-terrestrial missions, terrestrial analog datasets, and high-fidelity simulation environments, providing a critical foundation for validating autonomous mining algorithms under realistic constraints.
    \item \textbf{Challenges and future directions:} We delineate critical challenges and future trajectories across mechanical design, validation environments, and algorithmic architectures. Crucially, we highlight the transformative potential of Artificial Intelligence in augmenting robotic autonomy and decision-making for complex extraterrestrial operations.
    \item \textbf{Open-source project:} We have established an open-source platform as a centralized repository for space mining and robotics research. It synthesizes regulatory policies, industrial landscapes, academic literature, and international competitions, with a commitment to long-term curation and systematic updates. Our objective is to provide the global research community a dynamic gateway to the frontier of extraterrestrial resource development.
\end{itemize}

\begin{figure*}[t]
    \centering
    \includegraphics[width=\textwidth]{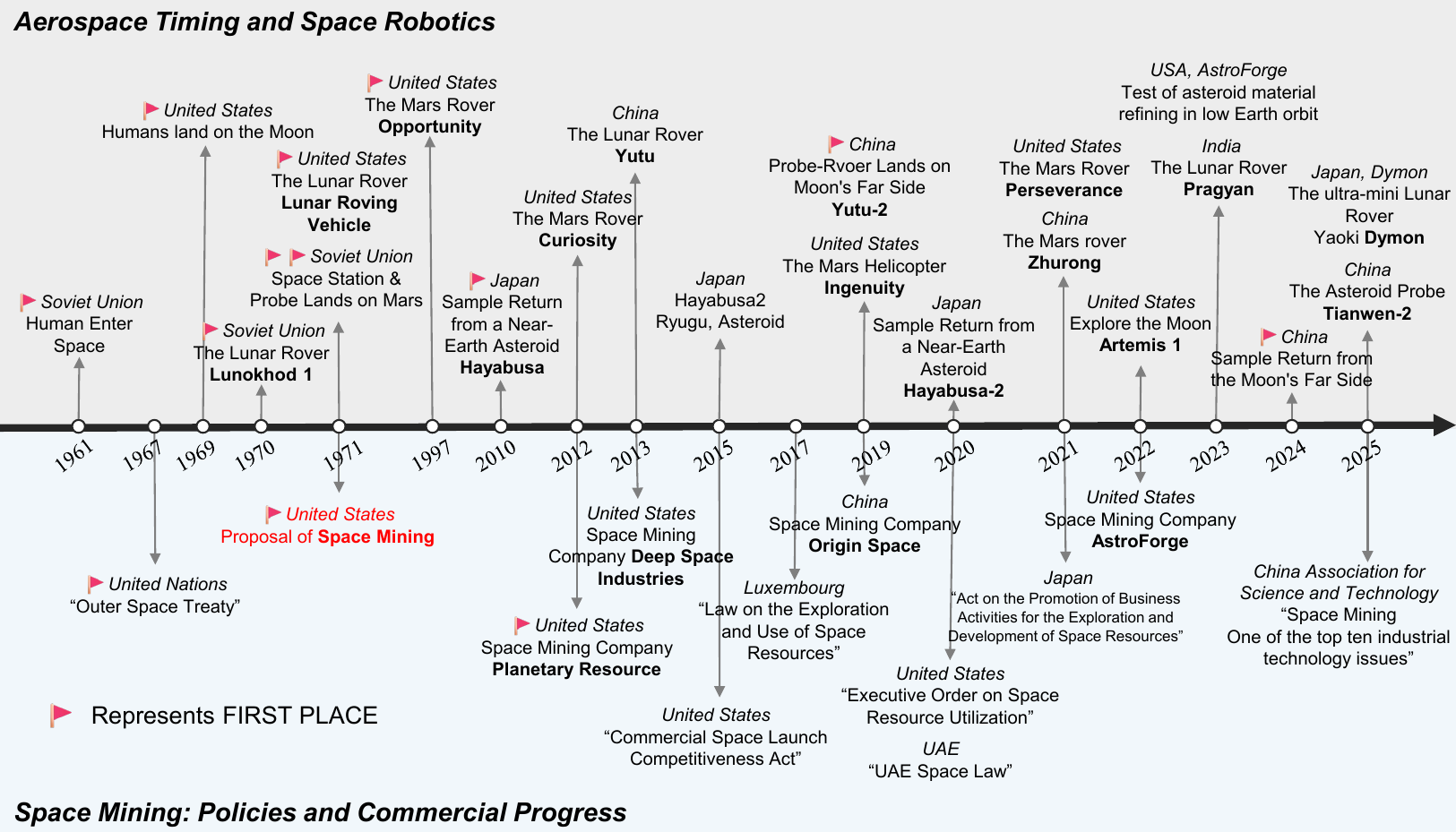}
    \vspace{-25pt} 
    \caption{\textbf{Timeline of Space Exploration and Mining Advancements.} Chronological overview highlighting major milestones in space exploration, key aerospace developments, and evolving international policies on space resource development. }
    \label{fig:timeline}
\end{figure*}

\section{Background} \label{sec:background}
\subsection{International Policies for Space Mining} \label{sec:policy}

The governance of space mining has transitioned from foundational international treaties to increasingly sophisticated national frameworks designed to catalyze large-scale commercial resource extraction. The 1967 United Nations \textit{Outer Space Treaty} remains the definitive cornerstone of this legal landscape, prohibiting national appropriation of celestial bodies while maintaining a strategic, permissive stance on private-sector utilization. This inherent ambiguity has prompted a wave of robust national legislation seeking to formally codify commercial rights. The United States led this shift with the 2015 \textit{Commercial Space Launch Competitiveness Act} and the 2020 \textit{Executive Order on Space Resource Utilization}. These legislative actions explicitly decoupled resource ownership from territorial sovereignty, a position further solidified by the U.S.-led Artemis Accords. Similarly, Luxembourg, the United Arab Emirates, and Japan have enacted legislation to grant legal title to extracted materials, reducing the systemic regulatory risks that once stifled private investment.

Parallel to these Western initiatives, China has identified extraterrestrial resource acquisition as a strategic priority. Following our proposal at the 2025 Annual Conference of the China Association for Science and Technology (CAST), space mining was officially designated one of the top ten industrial technology challenges, signaling a national commitment to establishing the commercial feasibility of autonomous operations. Collectively, these global policy shifts reflect a transition from theoretical exploration to a pragmatic race for resource sovereignty. By mitigating legal uncertainty and stimulating private capital, these frameworks underscore the necessity of autonomous systems as the primary agents of a compliant and sustainable multi-planetary economy. The chronological evolution of these milestones is illustrated in Figure~\ref{fig:timeline}.

\begin{table*}[ht]
\centering
\caption{\textbf{Key Industrial Entities in the Space Mining Sector.} Companies marked with an asterisk (*) are currently defunct.}
\vspace{10pt}
\label{tab:company}
\scriptsize
\begin{tabular}{lccl}
\hline
\textbf{Company} & \textbf{Founded} & \textbf{Location} & \textbf{Focus Areas} \\
\hline
Astrobotic Technology & 2007 & USA & Lunar robotics, payload delivery, ISRU demonstration rovers \\
Planetary Resources & 2009 & USA & Asteroid prospecting drones, robotic resource mapping \\
iSpace Technologies & 2010 & Japan & Lunar resource exploration, extraction of water and volatiles, robotic landers \\
Moon Express$^*$ & 2010 & USA & Robotic lunar mining and sample return missions \\
Deep Space Industries$^*$ & 2013 & USA & Micrometeorite capture satellites, in-space manufacturing and mining tech \\
Intuitive Machines & 2013 & USA & Lunar payload delivery, robotic landers, etc. \\
TransAstra & 2015 & USA & Optical mining systems for asteroids, robotic capture and resource utilization \\
Asteroid Mining Corporation & 2016 & UK & Robotic asteroid explorers, on-site extraction and transport \\
OffWorld & 2017 & USA & Autonomous mining robots for space and terrestrial critical minerals \\
Origin Space & 2019 & China & Asteroid resource prospecting and asteroid mining \\
Interlune & 2020 & USA & Lunar helium-3 harvesting, regolith excavation and processing, ISRU \\
Orbital Mining Corporation & 2022 & USA & Lunar regolith harvesting, robotic thermal processing for propellant and power \\
AstroForge & 2022 & USA & Asteroid mining, in-space resource processing, robotic refineries \\
\hline
\end{tabular}
\vspace{-20pt}
\end{table*}

\subsection{Commercial Space Mining Companies} \label{sec:company}

Since 2007, the commercial space mining sector has undergone a profound strategic evolution. Planetary Resources (2009) emerged as the first dedicated commercial entity to champion asteroid mining, establishing a visionary foundation for robotic extraterrestrial operations alongside early pioneers like Astrobotic Technology (2007) and ispace (2010). However, the early 2010s were also marked by the high-profile failure of conceptual ventures such as Moon Express and Deep Space Industries, whose ambitions in robotic sample return and micrometeorite capture proved premature.

A second, more resilient wave of industrial entities emerged as the broader commercial space economy matured. This includes TransAstra (2015), Asteroid Mining Corporation (2016), and OffWorld (2017), the latter developing autonomous robots for both terrestrial and space-based critical minerals. More recently, the sector has diversified with China's Origin Space (2019) and 2022 startups like AstroForge and Orbital Mining Corporation. This transition from the speculative "new concept" failures of the previous decade to today’s vibrant, multi-polar industrial race reflects the surge in commercial spaceflight capabilities. Yet, despite this commercial momentum, the industry remains at a critical juncture where long-term viability depends on overcoming the fundamental technical hurdles of autonomous execution in extreme environments.

\begin{table*}[ht]
\centering
\renewcommand{\arraystretch}{1.3} 
\caption{\textbf{Inventory of Deployed Extraterrestrial Robotic Explorers.} This survey catalogs mobile systems successfully operated on the Moon, Mars, and small bodies.}
\vspace{10pt}
\label{tab:robot}
\setlength{\tabcolsep}{3pt}
\scriptsize
\begin{tabular}{l c l l l p{7.2cm}}
\hline
\textbf{Robot} & \textbf{Nation} & \textbf{Target} & \textbf{Year} & \textbf{Duration} & \textbf{Key Features \& Strategic Strengths} \\
\hline
Lunokhod 1 & Soviet Union & Moon & 1970--71 & 11 mon. & First autonomous rover; 8WD solar-battery system. \\
Lunar Roving Vehicle & USA & Moon & 1971--72 & 11 hr. & Crewed electric mobility; foldable chassis for delivery. \\
Lunokhod 2 & Soviet Union & Moon & 1973 & 4 mon. & Improved 8WD suspension; extended surface traverse. \\
Yutu & China & Moon & 2013--16 & 31 mon. & Subsurface radar; thermal control for lunar night. \\
Yutu-2 & China & Moon (Far) & 2019-- & 6.8+ yr. & First far-side exploration; record mission longevity. \\
Pragyan & India & Moon & 2023 & 12 days & High-latitude analysis; detected elemental sulfur. \\
Sojourner & USA & Mars & 1997 & 85 days & First Mars rover; technology demo for solar mobility. \\
Spirit & USA & Mars & 2004--10 & 6 yr. & Rock abrasion tool (RAT); confirmed past water activity. \\
Opportunity & USA & Mars & 2004--18 & 15 yr. & 45-km distance record; aqueous geological evidence. \\
Curiosity & USA & Mars & 2012-- & 13+ yr. & Nuclear-powered (MMRTG); advanced analytical lab. \\
Perseverance & USA & Mars & 2021-- & 4+ yr. & Core sampling for return; biosignature reconnaissance. \\
Zhurong & China & Mars & 2021--22 & 1 yr. & First Chinese Mars rover; multi-spectral ground radar. \\
Jinchan & China & Moon (Far) & 2024 & Few days & 5-kg micro-rover; precision lander-docking and imaging. \\
Ingenuity & USA & Mars & 2021--24 & 3 yr. & First powered flight on Mars; 72 sorties; aerial scouting. \\
MASCOT & Germany/France & Ryugu & 2018 & 16 hr. & Asteroid micro-lander; swing-arm reorientation. \\
\hline
\end{tabular}
\end{table*}

\subsection{Planetary Rovers} \label{sec:rover}
To date, planetary surface mobility has been defined by discovery-driven exploration. Successfully deployed lunar and Martian rovers serve primarily as mobile remote laboratories for geological and environmental characterization. Flagship missions, such as NASA’s Perseverance and China’s Yutu-2, have demonstrated sophisticated autonomy in navigating extreme terrains; however, their strategic payloads are optimized for localized, small-scale analytical sampling rather than industrial-scale resource acquisition.
Perseverance exemplifies this high-precision scientific focus through its sophisticated suite of spectrometers and caching systems~\cite{verma2023autonomous,vicente2018seasonal}. Its Planetary Instrument for X-ray Lithochemistry (PIXL) and the Scanning Habitable Environments with Raman and Luminescence for Organics and Chemicals (SHERLOC) are designed to identify chemical elements and detect organic compounds at a microscopic scale. These instruments allow for the meticulous selection of rock cores, which are then hermetically sealed within specialized titanium tubes for future retrieval~\cite{bell2021mars,bhartia2021perseverance}. Similarly, Yutu-2, operating within the Von Kármán crater on the lunar far side, prioritizes deep-structure characterization. Its primary strategic payload, the Lunar Penetrating Radar (LPR), utilizes dual-frequency channels to map the subsurface stratigraphy and regolith thickness down to depths of several hundred meters. Combined with the Visible and Near-Infrared Imaging Spectrometer (VNIS), Yutu-2 provides unprecedented data on the mineralogical composition of the lunar mantle materials. These operations represent a pinnacle of remote sensing and passive geological surveying~\cite{wu2019lunar,ding20222,li2019chang}.

\begin{figure*}[htbp]
    \centering
    \includegraphics[width=0.9\textwidth]{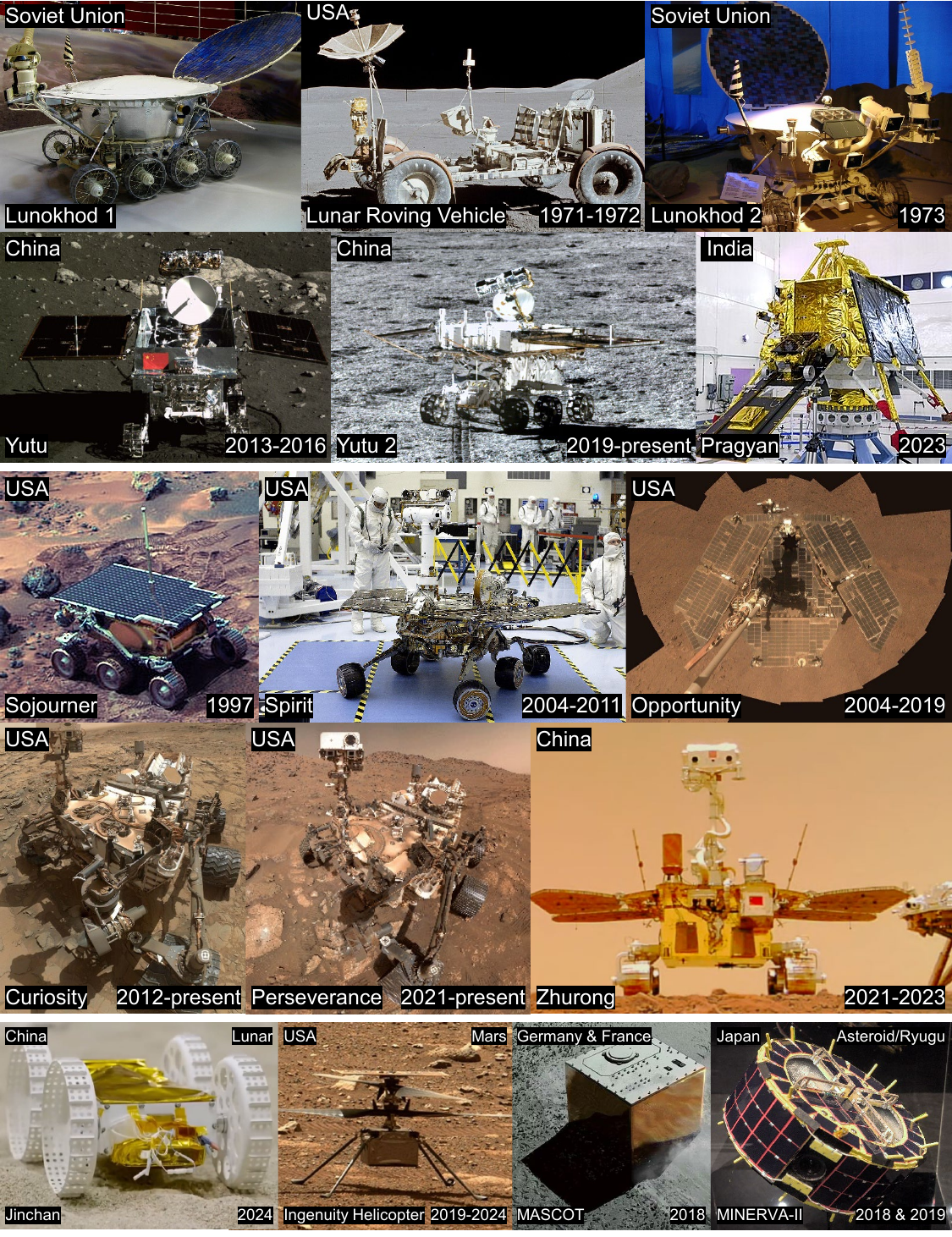}
    \vspace{-22pt} 
    \caption{\textbf{Trailblazers of Planetary Surface Exploration.} Montage of rovers and specialized robotic systems that have successfully operated on extraterrestrial surfaces. Lunar rovers: Lunokhod 1, Lunokhod 2, Yutu, Yutu-2, Pragyan. Mars rovers: Sojourner, Spirit, Opportunity, Curiosity,  Perseverance, Zhurong. Additional pioneering platforms: Jinchan (lunar sample-return manipulator), Ingenuity (Mars helicopter), and MASCOT (asteroid hopper on Ryugu). These systems represent key milestones in autonomous robotic exploration of the Moon, Mars, and small bodies.}
    \label{fig:robots_list}
\end{figure*}

\begin{table*}[ht]
\centering
\caption{\textbf{Selected Robotic Systems for Lunar and Martian Exploration and Resource Utilization (2026--2030).}}
\vspace{10pt}
\label{tab:rovers_future}
\scriptsize
\begin{tabular}{l l l l}
\hline
\textbf{Robot} & \textbf{Affiliation} & \textbf{Destination} & \textbf{Objective} \\
\hline
VIPER Rover & USA & Lunar & Map the location of water ice and other potential resources. \\
ISRU Pilot Excavator (IPEx) & USA & Lunar & Excavation and transport of lunar regolith for ISRU. \\
ATHLETE Hexapod & USA & Lunar & Heavy-duty cargo transport across lunar terrain. \\
Quadruped Hopping Robot & China & Lunar & Designed for low-gravity hopping or jet-assisted leaps. \\
Crewed Lunar Rover & China & Lunar & Pressurized crewed rover; currently in preliminary prototype development. \\
Sample Fetch Rover (SFR) & European & Mars & Retrieval of sample tubes cached by NASA's Perseverance rover. \\
Axel Two-Wheel Rover & USA & Lunar & Exploration of extreme terrain (craters, cliffs). \\
ExoMars rover & European & Mars & Search for evidence of past or present life on Mars. \\
\hline
\end{tabular}
\vspace{-10pt}
\end{table*}

As summarized in Table~\ref{tab:rovers_future}, the next generation of lunar and Martian rovers slated for deployment between 2026 and 2030 marks a decisive transition toward integrated resource prospecting and in-situ utilization. While earlier missions prioritized passive geological mapping, upcoming systems are engineered with explicit resource-oriented objectives. NASA’s VIPER~\cite{colaprete2021volatiles} and China’s Quadruped Hopping Robot exemplify this shift in prospecting strategy. VIPER is designed to map the concentration and distribution of water ice within permanently shadowed regions, whereas China’s hopping platform utilizes jet-assisted leaps to access extreme terrains previously unreachable by wheeled rovers. These systems serve as the vanguard for resource localization, providing the high-resolution spatial data necessary for subsequent extraction. The shift toward actual industrial operations is most prominently represented by the ISRU Pilot Excavator (IPEx). Unlike its predecessors that focused on milligram-scale sampling, IPEx is specifically designed for the high-volume excavation and transport of lunar regolith~\cite{schuler2022isru}. This platform serves as a critical technological bridge, transitioning from localized analysis to large-scale earthmoving operations in a vacuum environment. By demonstrating the capability to manipulate and move bulk quantities of abrasive regolith under low-gravity conditions, IPEx provides the essential operational foundation for autonomous space mining. Collectively, these advancements in adaptive manipulation and mission-specific mobility position the 2026--2030 cohort as the foundational enablers for sustained extraterrestrial industrialization and habitat construction.

\subsection{Space Mining and Terrestrial Open-Pit Mining} \label{sec:versus}
Space mining has emerged as a strategic frontier that represents the extraction of resources from asteroids, the Moon, and Mars to sustain deep-space exploration and seed future off-Earth industries. To accelerate this transition, it is essential to leverage the mature technological heritage of terrestrial mining. Historically, these Earth-based methods are categorized into open-pit, underground, and deep-sea operations. Among these, terrestrial open-pit mining offers the most compelling conceptual and operational parallel to extraterrestrial resource acquisition. The fundamental logic of surface-based excavation, which prioritizes the stripping of overburden to access extensive ore bodies, provides a robust baseline for designing the first generation of lunar and asteroidal mines~\cite{seweryn2024conceptual}.

The strategic alignment between terrestrial open-pit mining and space mining, particularly on the Moon and near-Earth asteroids, is defined by a shared reliance on surface-based material handling~\cite{dallas2020mining}. Both paradigms involve the systematic removal of surface layers, such as regolith or overburden, to reach target volatiles and minerals. However, profound environmental divergences necessitate a fundamental re-engineering of this legacy. While Earth-based open pits utilize massive mechanical energy and atmospheric stabilization, space operations must contend with microgravity, high-vacuum isolation, and extreme thermal cycling. These constraints dictate a shift from mass-intensive blasting and hauling toward precision-driven, low-reaction force excavation and stringent material containment. Such adaptations ensure that the open-pit methodology survives the transition to the lunar or asteroidal surface. The fundamental strategic and technological divergences between these two mining environments are summarized in Table~\ref{tab:versus}.

\begin{table*}[ht]
\centering
\caption{\textbf{Space Mining versus Terrestrial Open-Pit Mining}}
\vspace{10pt}
\label{tab:versus}
\renewcommand{\arraystretch}{1.3}
\setlength{\tabcolsep}{5pt}
\scriptsize
\begin{tabular}{@{} p{2.8cm} p{6.0cm} p{6.0cm} @{}}
\hline
\textbf{Dimension} & \textbf{Space Mining} & \textbf{Terrestrial Open-Pit Mining} \\
\hline
\textbf{Process} & Autonomous microgravity excavation; optical/microwave fragmentation; robotic drilling. & Large-scale blasting; bench-by-bench mechanical removal; heavy-duty loading. \\

\textbf{\begin{tabular}[c]{@{}l@{}}Equipment and \\ Technology\end{tabular}} & Lightweight, solar-powered rovers; integrated ISRU units; low-reaction drills. & Diesel-powered shovels and haul trucks; primary crushers; conveyor systems. \\

\textbf{Target Resources} & Volatiles (H$_2$O, He-3), structural metals (Ni, Fe), and PGMs for in-space use. & Bulk commodities (Fe, Cu, Coal) and industrial minerals for Earth-based supply chains. \\

\textbf{Primary Objective} & In-situ resource utilization (ISRU) to support propulsion, life-support, and construction. & Raw material supply for terrestrial industry and energy; maximizing throughput and yield. \\

\textbf{Primary Risks} & Vacuum, radiation, and thermal extremes; regolith containment; communication latency. & Geological instability; dust/water management; equipment wear; environmental impact. \\
\hline
\end{tabular}
\end{table*}

The cornerstone of this technological evolution lies in the transformation of autonomous systems. Terrestrial open-pit mining has already established a foundation through the deployment of large-scale autonomous haulage and drilling fleets~\cite{ge2022making, chen2024autonomous}. However, the move toward space mining requires a transition from these heavy-duty, earth-bound autonomous platforms to highly integrated, lightweight robotic swarms. Future lunar and asteroidal mines will not be powered by diesel-fueled giants but by agile, energy-efficient robotic units engineered for microgravity mobility and autonomous decision-making~\cite{zhang2025sustainable}. By evolving the core intelligence of terrestrial autonomous mining into the framework of lightweight space robotics, we can establish the essential operational infrastructure for sustained extraterrestrial industrialization.

\begin{figure*}[!t]
    \centering
    \includegraphics[width=\textwidth]{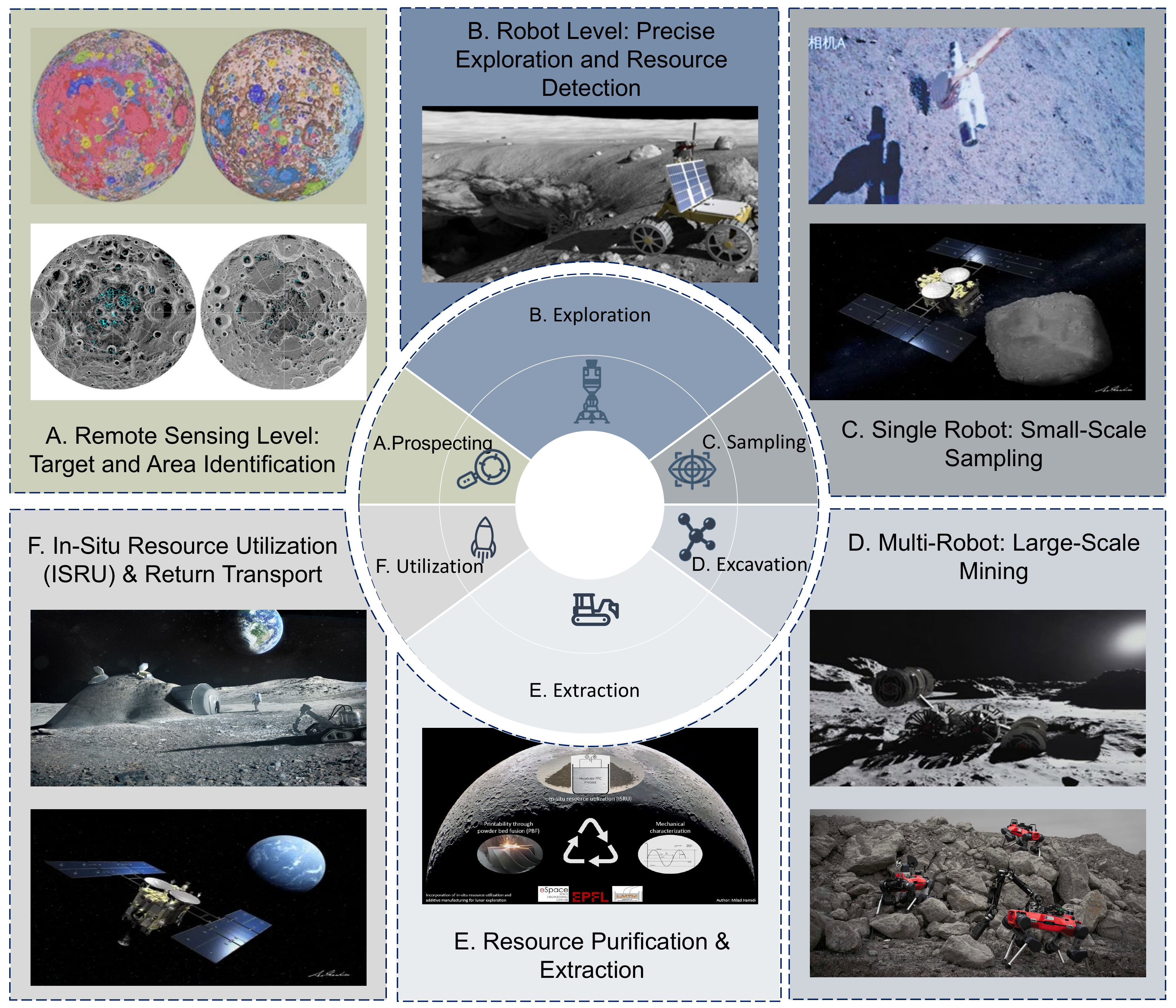}
    \vspace{-20pt} 
    \caption{\textbf{The Hierarchical Six-stage Framework for Autonomous Robotic Space Mining.} Schematic of the end-to-end operational lifecycle for extraterrestrial resource acquisition: (a) Prospecting for regional site selection; (b) Exploration via in situ robotic sensing; (c) Sampling for material validation; (d) Excavation for large-scale retrieval; (e) Extraction of refined volatiles or minerals; and (f) Utilization through in situ manufacturing or terrestrial transport.}
    \label{fig:pipeline}
\end{figure*}

\section{A Hierarchical Six-stage Framework for Space Mining} \label{sec:robotics}

We define a structured lifecycle for space mining that systematically bridges the gap between terrestrial engineering heritage and autonomous deep-space operations. This framework prioritizes a tripartite trajectory—comprising multi-scale exploration, autonomous sampling, and adaptive extraction—that unfolds through six discrete operational stages to ensure a robust pathway for off-Earth industrialization:
\begin{itemize}
    \item \textbf{(1) Prospecting: Remote sensing for target identification.} The process begins with macro-scale observation, leveraging orbital sensors and telescopic surveys to map celestial bodies. This stage identifies high-value volatiles and minerals, establishing the strategic baseline required for initial site selection.
    \item \textbf{(2) Exploration: Precise in situ robotic detection.} Transitioning to localized characterization, proximity spacecraft and mobile platforms conduct high-resolution analysis of surface composition and topography. This stage provides the granular data necessary for landing safety and mission-critical decision-making.  
    \item \textbf{(3) Sampling: Single-robot small-scale validation.} To validate material properties, autonomous robotic units conduct small-scale drilling and extraction tests. These operations verify the chemical quality and mechanical strength of the regolith under specific local environmental constraints.
    \item \textbf{(4) Excavation: Multi-robot large-scale harvesting.} Building upon initial validation, large-scale industrial systems or coordinated robotic swarms are deployed to harvest raw regolith. This stage scales the operational footprint, ensuring sustained material retrieval and high industrial throughput.
    \item \textbf{(5) Extraction: Resource purification and refinement.} Integrated systems process harvested materials to separate volatiles, metals, and water ice. These systems utilize task-adaptive manipulation to maintain efficient output and chemical purity within extreme extraterrestrial environments.
    \item \textbf{(6) Utilization: ISRU integration and return transport.} The final stage completes the operational loop by incorporating refined resources into In-Situ Resource Utilization (ISRU) frameworks. This supports off-Earth manufacturing and reduces dependence on terrestrial supply chains for long-term space industrialization.
\end{itemize}

\subsection{Stage 1: Remote sensing for strategic site selection and resource mapping}
Remote, non-contact detection of extraterrestrial resources has evolved from rudimentary global signatures to a sophisticated, multi-modal framework capable of delineating volatile-rich provinces and mineralogical heterogeneities at planetary scales. Early reconnaissance, conducted primarily through neutron, gamma-ray, and bistatic radar investigations, provided the foundational evidence for polar and near-surface hydrogen concentrations. The Clementine bistatic radar experiment identified anomalous polar backscatter suggestive of volatiles~\cite{Nozette1996}, while the Lunar Prospector neutron spectrometer generated the inaugural global geochemical blueprints of enhanced polar hydrogen, indicating the presence of buried water ice~\cite{Feldman1998}. These macro-scale indicators catalyzed targeted spectral and thermal follow-up missions. Specifically, hyperspectral data from Chandrayaan-1’s Moon Mineralogy Mapper (M\textsuperscript{3}) established widespread surficial hydration features~\cite{Pieters2009}, while the LCROSS impact and LRO/Diviner thermal mapping provided the definitive thermophysical context for polar cold traps~\cite{Colaprete2010, Paige2010}. Concurrently, advanced orbital radar, including LRO Mini-RF’s SAR observations, significantly refined the spatiotemporal constraints on the distribution and burial depth of sub-surface ice~\cite{Thomson2012, Patterson2017}.

Parallel advancements on Mars leveraged neutron and gamma-ray spectrometers aboard Mars Odyssey to quantify global near-surface hydrogen, guiding high-resolution mineralogical surveys. Visible–near-infrared (VNIR) hyperspectral instruments, such as OMEGA on Mars Express and CRISM on MRO, successfully mapped phyllosilicates, sulfates, and carbonates at hectometer-to-kilometer resolutions, effectively converting broad elemental anomalies into precise mineralogical targets~\cite{Boynton2004, Bibring2006, Murchie2007}. Furthermore, low-frequency sounding radars like MARSIS extended the detection horizon to kilometric depths, revealing high-dielectric subsurface reflectors beneath the south polar layered deposits—interpreted as potential brines or stable ice reservoirs~\cite{Orosei2018}.

For small bodies, rendezvous missions equipped with VNIR and thermal spectrometers (e.g., NEAR, Dawn, Hayabusa2, OSIRIS-REx) have demonstrated the robustness of remote sensing in detecting hydrated phases and organic compounds. Hayabusa2’s NIRS3 mapped distinct hydration features across Ryugu~\cite{Kitazato2019}, while OSIRIS-REx identified ubiquitous 2.7-$\mu$m absorption features on Bennu~\cite{Hamilton2019}. These findings confirm that orbital spectroscopy can provide reliable geochemical ground truth prior to physical interaction.

Synthesizing these modalities has established a standardized non-contact prospecting toolkit: neutron/gamma spectroscopy for bulk elemental proxies; radar sounding for subsurface dielectric profiling; VNIR–SWIR imaging for mineral characterization; and laser altimetry for precise topographic mapping. Current methodological trends emphasize the synergistic integration of these data streams to mitigate false positives and downscale orbital signals to mission-relevant spatial footprints. Representative examples include the joint analysis of neutron maps with thermal-spectral data~\cite{Feldman1998, Pieters2009, Paige2010} and multi-wavelength spectroscopy validated by returned samples~\cite{Kitazato2019, Hamilton2019}. Ultimately, this mature remote-sensing framework serves as the essential precursor for subsequent robotic In-Situ Resource Utilization (ISRU), delivering the high-fidelity maps required for autonomous landing and site-specific exploration.

\subsection{Stage 2: Precise in situ robotic detection}

While orbital remote sensing (Stage 1) provides essential macro-scale resource maps, its intrinsic resolution limits and indirect sensing modalities, often constrained by regolith burial depth and atmospheric interference, preclude the precise characterization required for industrial-scale acquisition. To bridge this gap, Stage 2 transitions to high-precision robotic detection, leveraging autonomous mobile platforms to provide definitive in-situ geological and structural ground truth. This robotic phase integrates four critical technology domains: (i) Robot Mechanisms and Motion Planning for hardware adaptability in extreme terrains; (ii) Localization and Mapping for spatial awareness in unknown, texture-sparse regions; (iii) Terrain Analysis and Navigation to address complex terramechanics during high-load traversal; and (iv) Sensors and Perception to maintain high-confidence situational awareness under harsh environmental conditions.

\textbf{Robot Mechanisms and Motion Planning}
The shift toward high-precision detection necessitates diverse robotic modalities specifically engineered to mitigate the environmental constraints of extraterrestrial bodies. Beyond traditional wheeled rovers, recent advancements prioritize mobility in three critical domains: (1) Rugged crater navigation, where legged systems leverage redundant degrees of freedom to traverse extreme inclines and rocky outcrops that are inaccessible to wheeled platforms~\cite{arm2023scientific, lee2025soft}; (2) Microgravity operations, where specialized anchoring mechanisms and multi-limbed grippers provide the necessary reaction force for drilling and instrument deployment on low-gravity asteroids~\cite{jiang2017robotic}; and (3) Soft-soil risk mitigation, involving task-adaptive motion planning to prevent wheel slippage and catastrophic vehicle entrapment in deep regolith or Martian sand~\cite{sutoh2012traveling, shrivastava2020material}.
In such scenarios, motion planning must transition from simple geometric obstacle avoidance to complex, terramechanics-aware manipulation. Robots must autonomously negotiate narrow lava tubes and unstable slopes while maintaining the dynamic stability required for sensitive geochemical analysis, ensuring mission continuity in environments where terrestrial mobility models inevitably fail.

\textbf{Localization and Mapping}
Extraterrestrial localization and mapping represent foundational challenges, providing the essential safeguards for autonomous deployment. The distinctive characteristics of these environments, particularly those of the Moon, Mars, and near-Earth asteroids, present three primary hurdles: (1) Unknown spatial priors, where the absence of dense satellite constellations, especially on the Moon, precludes high-fidelity prior maps, forcing missions to rely on simultaneous localization and mapping (SLAM) in unexplored terrains~\cite{wan2022terrain, sheppard2025marslgpr}; (2) Perceptual aliasing in weak-texture scenes, where the homogeneity of lunar regolith and Martian dust provides sparse visual landmarks, leading to drift in traditional odometry~\cite{tonasso2024lunar}; and (3) Harsh environmental variables, including pervasive dust-obscured visibility reminiscent of degraded visual environments in terrestrial mining, and extreme illumination contrasts in permanently shadowed regions (PSRs) where active sensing is mandatory~\cite{mahlknecht2022exploring}.

\textbf{Terrain Analysis and Navigation}
For heavy-duty mining and hauling rovers, navigation extends beyond geometry to complex terramechanics. (1) Regolith interaction: A critical challenge involves motion control in low-gravity, sandy terrains where non-linear wheel-soil interaction often leads to significant slippage or catastrophic vehicle entrapment~\cite{sutoh2012traveling}. (2) Hazardous topography: Extraterrestrial surfaces are characterized by extreme unevenness, ranging from impact-roughened lunar highlands to the chaotic boulder fields of asteroids. These features demand real-time terrain traversability analysis that accounts for both the physical dimensions of the robot and the load-bearing capacity of the soil~\cite{toupet2020terrain, xie2025mars, chen2025path}. Autonomous point-to-point navigation in these settings must therefore integrate reactive obstacle negotiation with long-term path optimality.

\textbf{Sensors and Perception}
Given the radiative and thermal extremes, sensor suites must be specifically ruggedized for extraterrestrial service. Standard optical cameras require advanced auto-exposure and high-dynamic-range (HDR) capabilities to mitigate the blinding glare of direct solar illumination in vacuum environments. Furthermore, to overcome the limitations of passive vision in dust-laden or light-devoid regions, specialized modalities, such as LiDAR for 3D structural mapping, Ground Penetrating Radar (GPR) for subsurface stratigraphy, and tactile sensors for direct regolith shear-strength assessment, are being integrated. These multisensor payloads ensure that robots can maintain high-confidence perception even when primary visual channels are compromised by Martian dust storms or lunar night.

\textbf{Robotic In-Situ Resource Exploration}
In planetary exploration, mobile platforms such as lunar and Martian rovers have evolved from conducting traditional scientific surveys to actively performing Robotic In-Situ Resource Exploration. The practical application of rover-based resource detection has already been demonstrated through the use of the Yutu rover to survey helium-3 distribution on the lunar surface ~\cite{ding2023moon}. Rather than relying solely on remote sensing, autonomous robotic systems for resource exploration enable localized, high-precision sampling and physical interaction with extraterrestrial regolith, with the objective of rapidly constructing detailed and dense resource distribution maps for potential resource targets ~\cite{li2026he3}. This paradigm shift is critical for identifying and quantifying key resources, particularly polar volatiles and water ice. To target these critical reserves, upcoming state-of-the-art missions such as China's Chang'e-7 mission and NASA's Volatiles Investigating Polar Exploration Rover (VIPER) are specifically engineered to traverse shadowed polar terrains for direct volatile prospecting.

\subsection{Stage 3: Single-robot small-scale sampling}

Small-scale sampling constitutes the critical empirical bridge between remote characterization and resource validation. Once orbital and robotic sensors delineate putative volatiles or mineral-rich provinces, dedicated acquisition systems perform direct regolith retrieval at centimeter-to-meter scales. This phase transitions mission objectives from predictive mapping to the definitive quantification of composition, stratigraphy, and extractability.

\textbf{Rotary–percussive Coring and Stratigraphic Integrity}
Coring systems provide stratified subsurface access while minimizing thermal and mechanical disturbances—a prerequisite for preserving volatile concentrations. State-of-the-art architectures, such as the Mars 2020 corer and the ESA ExoMars drill, integrate real-time torque sensing and autonomous feed-rate modulation to negotiate heterogeneous regolith profiles~\cite{nasacorer, Crawford2015LunarResources}. Furthermore, upcoming cryogenic prospectors, notably the TRIDENT drill on NASA’s VIPER mission, employ low-power actuation and dust-tolerant seals to retrieve intact cores from permanently shadowed regions, ensuring that the captured volatiles remain representative of their primary depositional state.

\textbf{Surface Abrasion and Contact-sensing Feedback}
For surficial characterization, robotic platforms utilize abrasion heads, scoops, and mechanical rasps to deliver sieved particulates to onboard analytical suites~\cite{Wiens2012ChemCam}. These tools operate within closed-loop frameworks, where contact spectrometers and microscopic imagers provide near-field feedback to refine sampling targets dynamically~\cite{Maurice2021SuperCam}. This iterative process ensures that high-value materials are prioritized for deeper extraction, optimizing the scientific yield under stringent mission energy constraints.

\textbf{Impulse-based Acquisition in Microgravity}
In the microgravity environments of near-Earth asteroids, sampling architectures rely on impulse-driven mechanisms to circumvent the requirement for complex anchoring forces~\cite{Lauretta2017OSIRISREx}. The TAGSAM system on OSIRIS-REx utilized gas-phase fluidization to mobilize regolith grains~\cite{Bierhaus2018TAGSAM}, while the Hayabusa2 sampling horn employed high-velocity projectile-induced ejecta capture~\cite{Grady2025RyuguOverview}. These rapid, "touch-and-go" maneuvers achieve gram-level collection within seconds, providing the physical ground truth necessary to calibrate remote spectroscopic observations of hydration and organic complexity~\cite{Hamilton2019BennuHydration}.

\textbf{Autonomous Adaptive Sampling and Target Selection}
Advanced autonomy increasingly governs the sampling lifecycle, mitigating the latency inherent in deep-space communications. Systems now synthesize multisensory data—including onboard computer vision, ground-penetrating radar, and thermal inertia maps—to optimize sampling sequences and safety margins~\cite{Martin2024AutonomousGuidance}. AI-driven frameworks, exemplified by the AEGIS system, enable autonomous target identification and rapid decision-making, which are essential for navigating the complex topographies of lunar polar or farside terrains~\cite{Estlin2012AEGIS}.

By establishing a direct correspondence between remote predictions and physical measurements, small-scale sampling provides the essential geochemical certification required for industrial-scale investment. This foundational stage is the precursor to integrated mining architectures, where autonomous decision-making loops synchronize in situ sampling with adaptive excavation to support pilot-scale In-Situ Resource Utilization (ISRU) and long-term extraterrestrial industrialization.

\subsection{Stage 4: Multi-Robot for Large-Scale Mining}

As planetary surface mining matures from conceptual frameworks to operational roadmaps, the technological trajectory has pivoted from monolithic, high-complexity excavators toward decentralized cooperative systems comprised of heterogeneous robotic swarms~\cite{cilliers2023toward, pirrone2025lunarSampling}. The extraterrestrial mining value chain, which encompasses prospecting, excavation, haulage, and In-Situ Resource Utilization (ISRU) processing, mandates operations that are both spatially distributed and temporally concurrent~\cite{zhang2023overview}. However, the combination of reduced gravity, extreme thermal cycling, abrasive regolith, and significant communication latency renders traditional single-agent teleoperation paradigms insufficient for sustained industrial throughput~\cite{sanders2005isruRoadmap, abel2023lunarDust}. Consequently, contemporary research advocates for multi-robot systems that decompose complex pipelines into specialized functional roles, utilizing distributed control to achieve parallelization and inherent fault tolerance across tasks ranging from subterranean skylight inspection to large-scale regolith harvesting~\cite{talamali2021whenLess, dominguez2025cooperativeSkylight, stonge2019planetaryTeams, rapp2024nearTermISPP, govindaraj2020multiRobotLunarBase}.

\textbf{Algorithmic Frameworks for Decentralized Coordination} At the methodological level, advancements in multi-agent coordination provide the robust algorithmic foundations necessary for persistent, long-duration missions~\cite{mcbeth2023scalable, matsiko2024overcoming, elromeh2025multirobot}. Motion-planning frameworks, particularly topology-guided planners designed for congested workspaces, have demonstrated significantly enhanced team-level efficiency within constrained planetary terrains~\cite{mcbeth2023scalable}. Furthermore, multi-objective and energy-aware coordination models facilitate scalable cooperation among heterogeneous fleets by balancing operational efficiency with energetic robustness. Complementing these are collective-intelligence models inspired by biological systems, which achieve decentralized coordination under limited sensing and communication constraints, thereby improving exploration coverage and resource-harvesting performance in high-fidelity simulations and terrestrial analogue tests~\cite{nitti2025collective, elromeh2025multirobot}.

\textbf{Operational Instantiations and System Integration} Building upon these theoretical foundations, several frameworks have instantiated multi-robot coordination in mission-specific lunar scenarios~\cite{chu2025lunarbase, elromeh2025multirobot, rocamora2023isru, zhang2025multimodal, luna2023emrs}. Recent developments include collaborative path-planning architectures for persistent base construction that integrate energy-aware scheduling with global motion planning under lunar environmental constraints~\cite{chu2025lunarbase}. Advanced multi-objective swarm algorithms have further optimized mapping performance in planetary-like environments by coupling coordinated coverage with adaptive exploration schemes~\cite{elromeh2025multirobot}. On the hardware-software interface, high-fidelity virtual environments now allow for the seamless integration of scouting, excavation, and haulage workflows~\cite{rocamora2023isru}. These efforts are mirrored by hardware advancements, such as multi-modal robotic platforms with tailored mechanisms for surface exploration and modular rover systems designed for cooperative payload handling in lunar-analogue testbeds~\cite{zhang2025multimodal, luna2023emrs}.

\textbf{Industrial Maturation and Commercial Testbeds} Engineering practice and commercial initiatives are providing the hardware platforms and test environments essential for advancing space-mining robotics~\cite{singh2022roboticlunar}. NASA’s Swamp Works has developed the RASSOR family of regolith excavators, which use counter-rotating drums to mitigate reaction forces under low gravity while demonstrating integrated excavation–haulage–feeding sequences~\cite{mueller2022extraterrestrial, gill2015rassor}. European efforts, including PERASPERA and PRO-ACT, have showcased multi-robot assembly of ISRU processing stations and mobile gantries, combining rovers with robotic arms for cooperative infrastructure construction~\cite{govindaraj2020multirobot}. Commercial players are accelerating progress: OffWorld deploys heterogeneous robotic fleets designed for swarm-based mining~\cite{frischauf2017offworld}, whereas Lunar Outpost's MAPP and Astrobotic's CubeRover adopt a modular architecture that pairs a common chassis with application-specific payloads for exploration and power services~\cite{gemer2024lunar}. GITAI has further demonstrated the integration of mining, construction, and maintenance functions using multi-arm mobile platforms for solar-panel deployment and module towing.

\textbf{Evolution Toward Heterogeneous and Modular Autonomy} The architecture of extraterrestrial multi-robot systems (MRS) has shifted from homogeneous configurations, designed primarily for redundancy and spatial coverage, to heterogeneous fleets that integrate specialized platforms for complementary sensing and manipulation~\cite{dominguez2025cooperative}. Greater sophistication is evident in modular, self-reconfigurable robots that can dynamically assemble into task-specific structures, enhancing adaptability in unpredictable terrains~\cite{dettmann2022corob}. These autonomous systems are increasingly embedded within human-robot interaction frameworks that support teleoperated oversight, shared autonomy, and collaborative decision-making, bridging Earth-based operators with in-situ robotic teams to sustain planetary industrialization~\cite{jusner2024mission, wang2025human, touma2025ai}.

\subsection{Stage 5: Resource Purification and Extraction}

The technological imperative for Stage 5 arises from the physical state of extraterrestrial resources, which are typically sequestered as trace components within bulk regolith. Helium-3, for instance, is trapped within the crystal lattices of lunar soil at parts-per-billion concentrations, necessitating intensive liberation processes to achieve usable volumes. Beyond the inherent form of these materials, the stringent purity requirements of mission-critical hardware, such as propulsion systems and life-support modules, mandate high-fidelity refining to eliminate abrasive particulates and chemical contaminants. By producing aerospace-grade consumables in situ, missions can circumvent the prohibitive logistical costs and mass constraints of Earth-to-orbit transport, thereby shifting the operational paradigm from mass-limited exploration to resource-independent industrialization~\cite{giel2025design}.

Methodologies for resource liberation are strictly dictated by the chemical and physical properties of the target species. For lunar volatiles and water ice, autonomous systems utilize thermal desorption within specialized reactors to release trapped gases, which are subsequently captured through cryogenic cold-trap condensation. In contrast, the extraction of oxygen and structural metals from oxide minerals relies on high-temperature chemical pathways, such as the hydrogen reduction of ilmenite or molten salt electrolysis. Recent multi-agent frameworks, exemplified by the work of Martinez et al., focus on synchronizing these energy-intensive thermochemical cycles with a continuous supply of raw materials provided by robotic excavation fleets~\cite{martinez2023multi}. For rare isotopes like Helium-3, extraction is further coupled with sophisticated isotopic separation techniques, ensuring the high-yield fuel quality necessary for future nuclear fusion architectures.

\subsection{Stage 6: In-Situ Resource Utilization (ISRU) or Return Transport}

The culmination of successful resource extraction lies in the transition toward autonomous additive manufacturing using local lunar or Martian regolith. Recent advancements in microwave sintering and binder jetting technologies facilitate the fabrication of structural components directly from minimally processed materials, which bypasses the prohibitive energy costs associated with transporting terrestrial bulk materials~\cite{suhaizan2024regolith,yashar2026planetary}. These in-situ construction paradigms enable the automated assembly of radiation-shielded habitats and launch infrastructure, providing the essential infrastructure for establishing permanent human outposts and sustained planetary presence~\cite{chu2025lunarbase}.

Beyond local infrastructure development, Stage 6 addresses the logistical framework required for returning high-value commodities to terrestrial markets. Helium-3 extracted from lunar regolith presents a potential fuel source for future fusion energy systems, whereas platinum-group metals derived from near-Earth asteroids could reinforce supply chains for Earth's clean-energy transition~\cite{pirrone2025lunarSampling}. Effective return transport requires the coordination of re-entry capsules with planetary launch windows, a synchronization that closes the economic loop of extraterrestrial mining and integrates space-derived materials into the global industrial economy~\cite{rios2024platinum}.

\section{Data for Space Mining} \label{sec:test}
Data are indispensable for operationalizing space mining and validating robotic architectures, yet they remain scarce relative to terrestrial repositories. These data facilitate performance benchmarking and the training of autonomous models for complex off-world missions. We evaluate current progress across three domains: datasets, simulation platforms, and terrestrial analogs.

\begin{table*}[ht]
\centering
\caption{\textbf{A Comprehensive List of Datasets for Extraterrestrial Scenarios.}}
\label{tab:slam_comparison}
\vspace{10pt}
\setlength{\tabcolsep}{1.5pt}
\scriptsize
\begin{tabular}{l l l l l p{2.5cm} l l p{2.8cm}}
\hline
\textbf{Dataset} & \textbf{Year} & \textbf{Env.} & \textbf{Type} & \textbf{Platforms} & \textbf{Sensors} & \textbf{GT} & \textbf{Tasks} & \textbf{Key Features} \\
\hline
DIND~\cite{furgale2012devon} & 2012 & Mars & Real & Wheeled & RGB, Sun sensor, Inclinometer & DGPS & Localize. & 1.5 km traverse \\
Canadian 3D Mapping~\cite{tong2013canadian} & 2013 & Mars & Real & Wheeled & Stereo, Laser, IMU & DGPS & SLAM & Heterogeneous platforms \\
Katwijk Beach~\cite{hewitt2018katwijk} & 2018 & Mars & Real & Wheeled & Stereo, ToF, LiDAR & Aerial & SLAM & Rock-strewn beach \\
LRU~\cite{vayugundla2018datasets} & 2018 & Mars & Real & Wheeled & Stereo, IMU, Odom & DGPS & Localize. & Rugged volcanic terrain \\
Labelmar~\cite{schwenzer2019labelmars} & 2019 & Mars & Real & Wheeled & RGB & - & Segment. & Authentic Martian data \\
Lunar Landscape~\cite{artificial_lunar_rocky_dataset}& 2019 & Lunar & Synth. & - & RGB & - & Segment. & Synth. from real data \\
Canadian Energy Aware~\cite{lamarre2020canadian} & 2020 & Mars & Real & Wheeled & Omni-Stereo, IMU, Pyranometer & GPS & SLAM & Solar radiation aware \\
Erfoud~\cite{lacroix2020erfoud} & 2020 & Mars & Real & Wheeled & LiDAR, Stereo & RTK & Localize. & Desert analog env. \\
Ai4mars~\cite{swan2021ai4mars} & 2021 & Mars & Real & Wheeled & RGB & - & Segment. & Crowd-sourced labels \\
MADMAX~\cite{meyer2021madmax} & 2021 & Mars & Real & Wheeled & LiDAR, RGB, IMU, Omni & RTK & SLAM & Planetary-analog \\
S3LI~\cite{giubilato2022challenges} & 2022 & M/Moon & Real & Wheeled & Stereo, LiDAR, IMU, GNSS & RTK & SLAM & Volcanic ash terrain \\
TAIL-Plus~\cite{wang2024we} & 2024 & Mars & Real & Wh/Quad & LiDAR, RGB, IMU & RTK & SLAM & Night-time operations \\
LuSNAR~\cite{liu2024lusnar} & 2025 & Lunar & Synth. & - & LiDAR, RGB, IMU & - & SLAM & Semantic labels \\
SPICE-HL3~\cite{rodriguez2025spice} & 2025 & Lunar & Real & Wheeled & Stereo, IMU, SPAD, Odom & OptiTrack & SLAM & Extreme lighting \\
Multi-Robot Mapping~\cite{lajoie2025multi} & 2025 & Mars & Real & Wheeled & LiDAR, IMU, Odom & RTK & SLAM & Collaborative mapping \\
\hline
\end{tabular}
\vspace{-4pt}
\begin{flushleft}
\scriptsize 
\textbf{Note:} \textbf{Env.}: Environment (M/Moon: Mars/Moon); \textbf{Type}: Data Type (Synth.: Synthetic); \textbf{GT}: Ground Truth; \textbf{Platforms}: Wh/Quad (Wheeled/Quadruped); \textbf{Sensors}: IMU (Inertial Measurement Unit), ToF (Time-of-Flight), GNSS (Global Navigation Satellite System), SPAD (Single-Photon Avalanche Diode), Odom (Wheel Odometry); \textbf{GPS}: DGPS (Differential GPS), RTK (Real-Time Kinematic); \textbf{Tasks}: Localize. (Localization), Segment. (Segmentation).
\end{flushleft}
\end{table*}

\subsection{Space Robot Datasets} \label{sec:datasets}

High-fidelity datasets are indispensable for the training of perception models and the preliminary validation of space robotics. From a methodological perspective, these data sources are classified into a tripartite framework. First, legacy datasets collected by Martian rovers, lunar rovers, or Apollo astronauts during early exploration represent the only direct ground truth, yet they are characterized by extreme scarcity and frequent quality discontinuities~\cite{swan2021ai4mars, schwenzer2019labelmars}. Second, synthetic datasets generated within high-fidelity simulation environments provide precise pose ground truth and comprehensive semantic labeling, although they are often compromised by limited rendering fidelity and a lack of photorealism~\cite{liu2024lusnar}. Third, datasets acquired in terrestrial analog environments offer extensive multimodal sensor integration and diverse environmental stressors, including direct solar illumination and pervasive dust, but they remain fundamentally constrained by the inherent domain gap between Earth-based and extraterrestrial physics~\cite{rodriguez2025spice, tong2013canadian, cloutis2015canadian, lamarre2020canadian, lajoie2025multi}.

To address these limitations, recent advancements in world models represent a scalable frontier in dataset synthesis~\cite{li2025martian}. These architectures can generate entire environmental sequences from limited single-view images, offering a robust methodology for creating large-scale, high-fidelity datasets that accurately reflect the complex conditions of space environments. By bridging the sim-to-real gap through learned temporal and spatial dynamics, such generative approaches are becoming essential for the development of next-generation autonomous explorers.

\begin{table*}[ht]
\centering
\caption{\textbf{Overview of Robotic Simulators for Extraterrestrial Environments.} Asterisks (*) denote proprietary or closed-source systems.}
\vspace{10pt}
\label{tab:simulators}
\scriptsize
\begin{tabular}{l c c c c c}
\hline
\textbf{Simulator} & \textbf{Years} & \textbf{Scenes} & \textbf{Renderer} & \textbf{Robots} & \textbf{Tasks} \\
\hline
RP Simulator~\cite{allan2019planetary}* & 2019 & Lunar & ROS Gazebo & Wheeled & Navigation, Localization \\
MarsSim~\cite{giubilato2020simulation} & 2020 & Mars & ROS Gazebo & Wheeled & Perception, Localization \\
Space Robotics Challenge* & 2021 & Lunar & ROS Gazebo & Wheeled & Lunar Mining, Navigation, etc. \\
MarsSim~\cite{zhou2022marssim}* & 2022 & Mars & ROS Gazebo & Wheeled & Perception, Navigation \\
LunarSim~\cite{pieczynski2023lunarsim} & 2023 & Lunar & ROS Gazebo & Wheeled & Segmentation, Localization, Navigation \\
Lunar Autonomy Challenge* & 2024 & Lunar & CARLA & Wheeled & Perception, Mapping, Localization \\
RLROVERLAB~\cite{mortensen2024rlroverlab} & 2024 & Lunar, Mars & Isaac Sim & Wheeled & Navigation, Grasping \\
OmniLRS~\cite{richard2024omnilrs} & 2025 & Lunar & Isaac Sim & Wheeled & Segmentation, Navigation, Mapping \\
Space Robotics Bench~\cite{orsula2025space} & 2025 & \begin{tabular}[c]{@{}l@{}}Orbit, Moon, \\ Mars, Asteroid\end{tabular} & Isaac Sim & \begin{tabular}[c]{@{}l@{}}Wheeled, Legged, \\ Aerial, Spacecraft\end{tabular} & Landing, Navigation, Excavation, etc. \\
\hline
\end{tabular}
\end{table*}

\subsection{Space Scene Simulators}  \label{sec:simulator}
Simulators serve as indispensable platforms for the rapid prototyping of extraterrestrial mining systems while providing the high-volume datasets necessitated by the training of advanced autonomous models. Initial simulation frameworks, designed primarily for the real-time testing of navigation and exploration protocols, were largely established on the Robot Operating System (ROS) integrated Gazebo environment, as exemplified by RP Simulator and LunarSim~\cite{allan2019planetary, pieczynski2023lunarsim}. The principal advantage of these systems lies in their native ROS compatibility, which affords researchers access to a sophisticated ecosystem of established robotics tools and libraries. However, Gazebo frequently exhibits significant fidelity gaps in rendering and physics modeling, particularly regarding fluid dynamics, regolith interactions, and complex environmental backgrounds. This framework was notably utilized by NASA to develop a lunar simulation environment for space-mining competitions, which required the implementation of algorithms for localization, mapping, navigation, and multi-robot coordination.

Driven by realism requirements, subsequent studies transitioned to AirSim and CARLA for planetary platforms~\cite{shah2017airsim, dosovitskiy2017carla, tian2025slam}. Currently, the prevailing methodology centers on NVIDIA Isaac Sim, which is favored for its seamless ROS integration, advanced physics engine, and photorealistic rendering capabilities~\cite{orsula2025space, richard2024omnilrs, mortensen2024rlroverlab}. Furthermore, its mature interfaces facilitate robust sim-to-real pipelines and reinforcement learning architectures, thereby bridging the gap between algorithmic development and physical deployment~\cite{mittal2023orbit, isaaclab2025}. A comprehensive summary of prominent space robotic simulators is provided in Table~\ref{tab:simulators}.

\begin{figure*}[!t]
    \centering
    \includegraphics[width=0.8\textwidth]{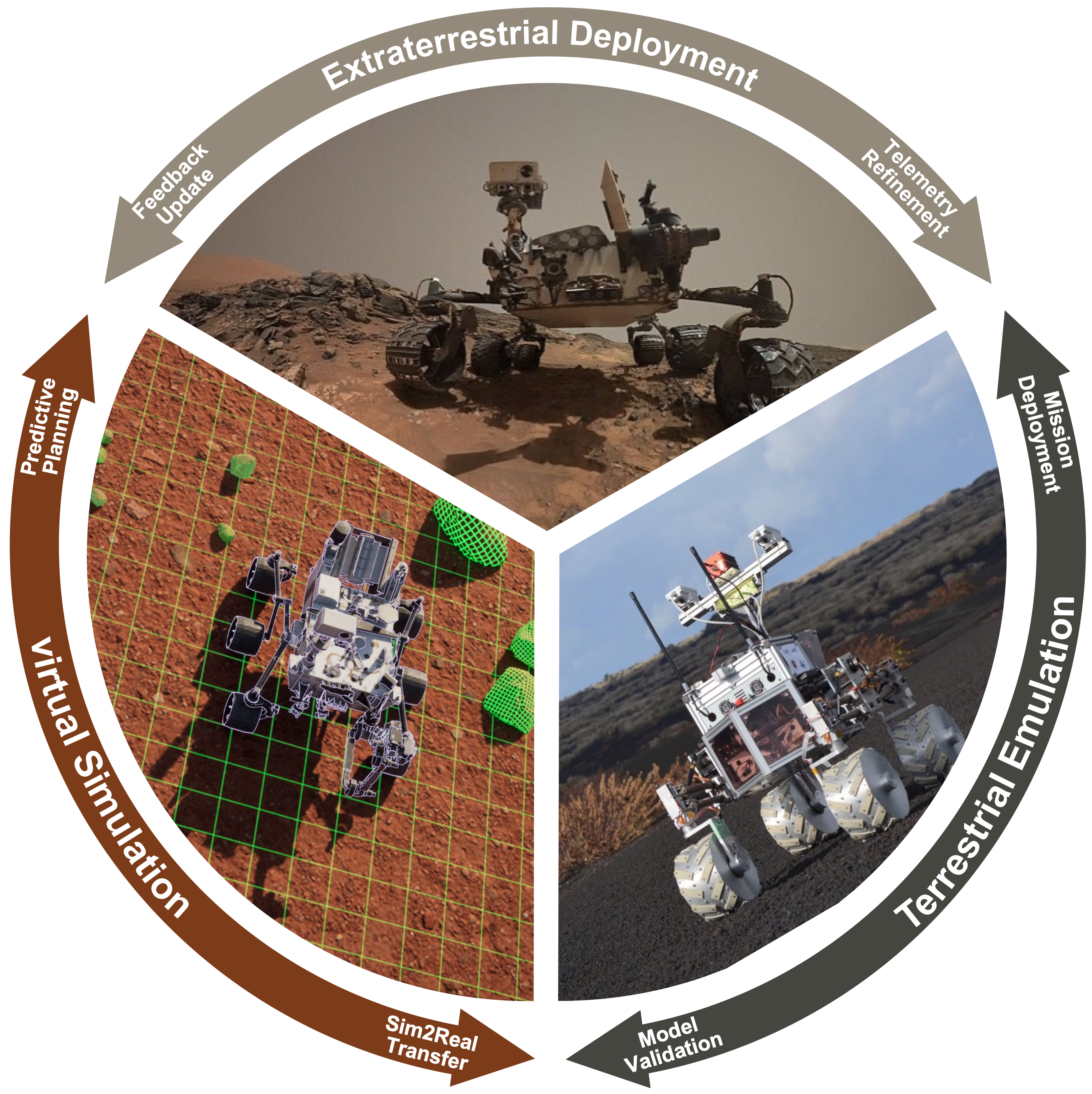}
    \vspace{-15pt} 
    \caption{\textbf{Closed-loop Data Cycle for Space Robotics.} From simulation testing to terrestrial emulation validation, on to real-world deployment, with bidirectional enhancement of both simulation environments and terrestrial emulators. }
    \label{fig:data}
\end{figure*}

\subsection{Earth-based Analogs of Extraterrestrial Environments}

Terrestrial simulation of extraterrestrial environments is indispensable for validating lunar and Martian rover architectures before deployment, yet such analogs remain fundamentally limited by the inescapable physical constraints of Earth. Replicating microgravity, which constitutes a defining operational parameter for missions on the Moon (approximately 1/6 g) and Mars (approximately 3/8 g), represents a persistent challenge. Conventional methodologies, including parabolic flights, drop towers, neutral buoyancy facilities, and cable-suspended offloading, generally afford reduced gravity for only transient durations. Furthermore, these techniques typically apply localized compensation forces to a vehicle’s center of mass or discrete contact points, failing to achieve the uniform force distribution required at the critical wheel-regolith interface~\cite{yoshimitsu1999microgravity, yoshimitsu2004development, nakamura2000mobility}.

While vacuum chambers successfully emulate atmospheric absence and extreme thermal gradients, and high-fidelity regolith simulants, such as JSC-1A or EAC-1A, facilitate realistic surface-mechanics modeling, the lack of sustained, whole-body microgravity hinders the dynamic assessment of gait adaptation, traction control, and disturbance rejection. These capabilities are pivotal for resource-intensive operations, particularly excavation and precision sample manipulation. Consequently, physical analog testing must be synergistically integrated with high-fidelity synthetic environments and targeted hardware-in-the-loop validation to address the developmental requirements of next-generation mining rovers.

\section{Challenges and Future Directions} \label{sec:directions}
Advancing extraterrestrial mining necessitates the seamless integration of physical capability, algorithmic intelligence, and empirical validation. These pillars collectively govern robotic agency and adaptation, which are critical for robust autonomy within extreme, data-sparse environments. However, contemporary rovers remain constrained by teleoperation, as communication latencies and environmental stochasticity impede operational throughput. Productive exploitation thus demands embodied geological intelligence, facilitating in-situ multimodal interpretation, material property estimation, and closed-loop decision-making from hypothesis to execution.

\subsection{Robotic Resource Acquisition in Microgravity} \label{sec:humanoid_robot}

\textbf{Challenges.} Microgravity diminishes tractive effort, as every interaction force generates a reactive impulse that compromises contact stability. Consequently, conventional locomotion and manipulation lose the restorative effect of gravitational weight, rendering wheels and tracks unable to generate sufficient normal force for robust engagement~\cite{pirrone2025lunarSampling}. Beyond kinematics, regolith in vacuum exhibits cohesion and electrostatic activity, which exacerbate tool wear, clog conveyors, and induce fluidization during excavation~\cite{Qi2025CohesiveLunarSoil}. Volatile-rich materials impose further stressors on sealing and thermal management. Hardware must withstand extreme thermal cycling, radiation, and pervasive dust while maintaining high structural stiffness, necessitating sensors capable of resolving minute interaction forces without saturation or backdriving.

\textbf{Future directions.} Advancement centers on momentum-aware, multimodal architectures designed to decouple surface interaction from gravitational dependence. Robotic platforms should integrate heterogeneous mobility solutions, including microspines for irregular rock, gecko-inspired adhesives for smooth surfaces, and compact anchoring mechanisms for high-torque drilling. Distributing these mechanisms across multi-limbed systems, combined with reaction-aware gait planning, ensures stable attachment during high-load operations~\cite{Nadan2024LORIS}. Tooling strategies, such as reaction-cancellation via counter-rotating drives or symmetric dual-arm manipulation, effectively mitigate net disturbances during coring. Furthermore, gas-assisted sampling heads and well-sealed, dust-resistant containers reduce mechanical friction while preserving the integrity of sensitive components. Integrating carrier-daughter architectures and small hopping scouts further enhances system robustness, shifting the paradigm from single-purpose machines toward reconfigurable, scalable platforms for microgravity environments.

\subsection{Embodied Foundation Models for Space Mining} \label{sec:mm_llms}

\textbf{Challenges.} Despite significant advances in perception and navigation, contemporary space robotics remains heavily tethered to ground operators due to communication latencies, restricted situational awareness, and insufficient robustness in contact-intensive maneuvers~\cite{verma2023autonomous}. This dependence severely diminishes operational throughput. Autonomous mining presents fundamentally greater complexities than traversal, as it necessitates the coupling of perception with forceful interaction while simultaneously reasoning about stochastic material properties~\cite{Louca2025RegolithTeleop}.
While terrain mapping and obstacle detection are relatively mature, geological interpretation and physical property estimation remain predominantly ground-based~\cite{Chien2024SpaceRoboticsReview}. Multimodal scientific datasets, such as reflectance spectra and ground-penetrating radar, are typically downlinked for offline analysis, thereby delaying critical decisions regarding excavation and real-time adaptation. In microgravity environments, interaction dynamics further constrain feasible behaviors, often rendering high-level plans unexecutable without grounding in complex contact constraints. Furthermore, data scarcity, domain shifts across regolith types, and limited onboard computation hinder the development of robust multimodal representations and verifiable safety protocols~\cite{Dettmers2023QLoRA}.

\textbf{Future directions.} Future systems must transition from scripted protocols toward embodied geological intelligence, which integrates multimodal sensing with interaction feedback such as force, torque, and contact stability. Such integration facilitates the real-time inference of material states, including cohesion, bearing strength, and volatile indicators, to guide autonomous excavation and sampling~\cite{pirrone2025lunarSampling}. A viable framework involves hierarchical architectures that combine foundation models with structured skill libraries and physics-aware critics, ensuring that generated plans remain consistent with robot dynamics and tool constraints.
World models that merge geometric-semantic mapping with predictive material responses can support active perception and long-horizon task planning through predictive pre-action evaluation~\cite{Chien2024SpaceRoboticsReview}. Operational safety can be reinforced through pre-execution feasibility checks, uncertainty-aware gating, and lightweight runtime safeguards that enable transparent failure detection. Given constrained onboard resources, parameter-efficient adaptation and continual self-supervised learning are essential for maintaining robustness without frequent retraining~\cite{Dettmers2023QLoRA, HafezWermter2024ContinualRobotLearning}. Finally, embedding geological priors and instrument ontologies into prompts and state representations will enhance onboard reasoning while preserving system auditability.

\subsection{Simulation and Validation for Space Mining Autonomy} \label{sec:simulated_env}

\textbf{Challenges.} Faithful replication of space mining conditions remains difficult because microgravity, vacuum, thermal cycling, radiation, and regolith behaviors such as cohesion, electrostatics, and fluidization are rarely reproduced together. Gravity offload systems, neutral buoyancy, and vacuum chambers each introduce scale effects, boundary artifacts, or contact distortion. Dust transport, illumination variability, and long-term degradation are also commonly simplified \cite{pirrone2025lunarSampling}. Fully virtual simulation struggles with contact-intensive excavation and regolith tool interaction, which leads to uncertain predictions of reaction impulses and throughput \cite{Louca2025RegolithTeleop}. In addition, datasets from simulation, terrestrial emulation, and limited in situ trials often differ in metadata conventions and time synchronization, which hinders systematic calibration and transfer across platforms \cite{Chien2024SpaceRoboticsReview}. The absence of standardized tasks, metrics, and parameter identification pipelines further weakens validation and slows iterative progress.

\textbf{Future directions.} Future work should establish a closed-loop validation ecosystem that links simulation, terrestrial emulation, and extraterrestrial deployment. High-fidelity physics simulation should be developed alongside sensor simulation, including scientific instruments when relevant, to support policy training, uncertainty estimation, and exploration of rare failure modes. Digital twins can be maintained through automated parameter identification using contact and system telemetry, enabling continuous calibration of regolith tool interaction models and reaction dynamics. World models for embodied autonomy can be trained as hybrid predictors that combine physics-based components with learned residual terms, with periodic updates driven by emulation and in situ data. Hardware in the loop evaluation and standardized benchmarks, including contact stability, reaction impulse, excavation throughput, sampling integrity, and fault rates, can support consistent comparison across systems and scales \cite{Chien2024SpaceRoboticsReview}. This pipeline can convert isolated test campaigns into an adaptive process in which simulation generates candidates, emulation evaluates them, and deployment provides corrective evidence, improving reliability and autonomy for space mining.

\section{Conclusion}

Space mining represents a strategic imperative for the future of humanity, marking the inevitable transition from planetary confinement toward the broader exploration of the solar system. This endeavor addresses two critical dimensions of global sustainability: first, it provides a resilient alternative to the looming resource crises on Earth by accessing extraterrestrial material wealth; second, it serves as the primary energetic and material driver for deep-space missions. By facilitating the local production of propellants and structural feedstocks, space mining eliminates the logistical burden of Earth-to-orbit transport, thereby providing the foundational infrastructure necessary for establishing permanent human outposts and sustained planetary habitation.

As the ultimate culmination of modern industrial integration, space mining demands the seamless convergence of advanced robotics, autonomous systems, and materials science. Our proposed six-stage framework delineates the rigorous technical trajectory required for operational success, yet each phase presents formidable challenges that must be addressed to ensure mission viability. From the complexities of high-fidelity prospecting and robotic excavation in microgravity to the intricate requirements of in-situ purification and logistical return, these stages necessitate unprecedented levels of operational robustness. Overcoming the barriers of extreme environmental stressors, data scarcity, and complex interaction dynamics remains a fundamental prerequisite for transforming space mining from a conceptual framework into a reliable, revenue-generating pillar of the interplanetary economy.


\clearpage

\bibliography{main} 
\bibliographystyle{sciencemag}

\newpage

\renewcommand{\thefigure}{S\arabic{figure}}
\renewcommand{\thetable}{S\arabic{table}}
\renewcommand{\theequation}{S\arabic{equation}}
\renewcommand{\thepage}{S\arabic{page}}
\setcounter{figure}{0}
\setcounter{table}{0}
\setcounter{equation}{0}
\setcounter{page}{1}

\end{document}